\documentclass[letterpaper]{article} % DO NOT CHANGE THIS
\usepackage[]{aaai24}  % DO NOT CHANGE THIS
\usepackage{times}  % DO NOT CHANGE THIS
\usepackage{helvet}  % DO NOT CHANGE THIS
\usepackage{courier}  % DO NOT CHANGE THIS
\usepackage[hyphens]{url}  % DO NOT CHANGE THIS
\usepackage{graphicx} % DO NOT CHANGE THIS
\usepackage{natbib}  % DO NOT CHANGE THIS AND DO NOT ADD ANY OPTIONS TO IT
\usepackage{caption} % DO NOT CHANGE THIS AND DO NOT ADD ANY OPTIONS TO IT
\usepackage{algorithm}
\usepackage{newfloat}
\usepackage{listings}
\DeclareCaptionStyle{ruled}{labelfont=normalfont,labelsep=colon,strut=off} % DO NOT CHANGE THIS
\floatstyle{ruled}
\newfloat{listing}{tb}{lst}{}
\floatname{listing}{Listing}
\usepackage{amssymb,amsmath,amsfonts}
\usepackage{multicol,multirow,enumerate,textcomp} % setspace FORBIDDEN
\usepackage{stackengine,calc,xspace,array}
\usepackage{verbatim}

\DeclareSymbolFont{rsfscript}{OMS}{rsfs}{m}{n}
\DeclareSymbolFontAlphabet{\mathrsfs}{rsfscript}

\usepackage[dvipsnames]{xcolor}

\usepackage[makeroom]{cancel}

\usepackage{algorithm,algorithmicx}
\usepackage[noend]{algpseudocode}
\MakeRobust{\Call}
\usepackage[T1]{fontenc}

\title{Fast and Knowledge-Free Deep Learning for General Game Playing \\(Student Abstract)}
\author{
    Micha{\l} Maras, Micha{\l} K\k{e}pa, Jakub Kowalski, Marek Szyku{\l}a,
}
\affiliations{
    University of Wroc{\l}aw, Faculty of Mathematics and Computer Science\\
    mmaras1999@gmail.com, kempus1999@gmail.com, jko@cs.uni.wroc.pl, msz@cs.uni.wroc.pl
}

\begin{document}\maketitle\begin{abstract}
We develop a method of adapting the AlphaZero model to General Game Playing (GGP) that focuses on faster model generation and requires less knowledge to be extracted from the game rules. The dataset generation uses MCTS playing instead of self-play; only the value network is used, and attention layers replace the convolutional ones. This allows us to abandon any assumptions about the action space and board topology. We implement the method within the Regular Boardgames GGP system and show that we can build models outperforming the UCT baseline for most games efficiently. 
\end{abstract}

\section{Introduction}

General Game Playing (GGP) is an Artificial Intelligence challenge focused on developing an autonomous game-playing agent that can play, without human intervention, any game given its rules \cite{Genesereth2005General}.
Such a task requires a proper formalism to encode a possibly large class of games in a machine-processable and simultaneously human-readable way.
Several such formalisms were developed, Stanford's Game Description Language (GDL) \cite{Genesereth2005General} being arguably the most famous and deep-researched, providing a great domain for testing AI algorithms, especially Monte Carlo Tree Search (MCTS) with UCT \cite{Browne2012ASurvey}. 
A newer approach, focusing on more efficient game processing, is Regular Boardgames (RBG) \cite{Kowalski2019RegularBoardgames}, which encodes game rules as regular expressions. 
A natural path of GGP research is to apply Neural Networks (NN) and Deep Reinforcement Learning (DRL).

The most famous NN-based approach that was advertised as general was AlphaZero \cite{silver2018general}, applying the same learning algorithm for Go, Chess, and Shogi. However, although the proposed method really generalizes (at least among two-player, zero-sum board games), the network architecture had to be manually prepared for each game, which is inconsistent with the pure GGP principles.
A clone of AlphaZero, based on GDL and aimed to resolve some of these restrictions, was presented in \cite{ThielscherAAAI20}.
MCTS-based DRL approach for Ludii GGP system, created via a bridge to Polygames can be found in \cite{soemers2021deep}. 
Deep Reinforcement Learning using only value networks combined with a variant of Unbounded Minimax is described in \cite{cohen2023minimax}. 

We address some issues that appear in the GGP context and focus on training within a short time limit and further restricting assumptions about the given game rules.

\begin{table*}[!ht]\small\renewcommand{\arraystretch}{1.045}
\newcommand{\cb}[1]{{\scriptsize $\pm$#1}}
\newcommand{\oppA}{vs.\ $1\times$}
\newcommand{\oppB}{vs.\ $10\times$}
\begin{center}\begin{tabular}{|lrc|r|r|r|r|r|r|r|r|}\hline
\multicolumn{1}{|c}{\multirow{3}{*}{{\bf Game}}} & \multicolumn{2}{l|}{\multirow{3}{*}{\bf Board size}} & \multicolumn{4}{c|}{\bf Ordered board} & \multicolumn{4}{c|}{\bf Permuted board} \\\cline{4-11}
& & & \multicolumn{2}{c|}{{\bf Attention NN}} & \multicolumn{2}{c|}{{\bf Convolutional NN}} & \multicolumn{2}{c|}{{\bf Attention NN}} & \multicolumn{2}{c|}{{\bf Convolutional NN}} \\%\rule{0pt}{9pt}
\multicolumn{3}{|c|}{} & \oppA & \oppB & \oppA & \oppB & \oppA & \oppB & \oppA & \oppB \\\hline

Breakthrough & 36 & $(6\times6)$  & {99}\%\cb{1.1} & 94\%\cb{2.4} & {99}\%\cb{0.8} & \textbf{98}\%\cb{1.4} & \textbf{99}\%\cb{0.7} & {93}\%\cb{2.5}  & 97\%\cb{1.7} & {93}\%\cb{2.5} \\ %\cline{2-10}
Breakthrough & 64 & $ (8\times8)$ & 83\%\cb{3.7} & 45\%\cb{4.9} & \textbf{88}\%\cb{3.2} & \textbf{65}\%\cb{4.7} & \textbf{85}\%\cb{3.5} & \textbf{53}\%\cb{4.9} & 77\%\cb{4.1} & 42\%\cb{4.8} \\ %\cline{2-10}
Breakthrough & 100 & $(10\times10)$ & {51}\%\cb{4.9} & 9\%\cb{2.8} & {47}\%\cb{4.9} & \textbf{15}\%\cb{3.5} & \textbf{17}\%\cb{3.7} & \textbf{4}\%\cb{1.9} & 4\%\cb{2.0} & 1\%\cb{0.8} \\

\multirow{1}{*}{Connect Four} & 42 & $(7\times6)$  & \textbf{72}\%\cb{4.3} & {57}\%\cb{4.8} & 58\%\cb{4.8} & {53}\%\cb{4.8} & 44\%\cb{4.8} & 37\%\cb{4.6}  & \textbf{49}\%\cb{4.9} & \textbf{42}\%\cb{4.8} \\

\multirow{1}{*}{English Draughts} & 32 & $(8\times8)$  & {86}\%\cb{2.3} & {43}\%\cb{3.3} & {85}\%\cb{2.5} & {46}\%\cb{3.4} & 77\%\cb{2.8} & 36\%\cb{3.1}  & \textbf{80}\%\cb{2.7} & \textbf{43}\%\cb{3.3} \\

\multirow{1}{*}{Fox and Hounds} & 32 & $(8\times8)$  & {91}\%\cb{2.8} & \textbf{77}\%\cb{4.1} & {93}\%\cb{2.5} & 72\%\cb{4.4} & 87\%\cb{3.3} & 58\%\cb{4.8}  & \textbf{94}\%\cb{2.4} & \textbf{72}\%\cb{4.4} \\

\multirow{1}{*}{Gomoku} & 225 & $(15\times15)$  & \textbf{88}\%\cb{3.2} & {38}\%\cb{4.8} & 44\%\cb{4.9} & {39}\%\cb{4.8} & \textbf{74}\%\cb{4.3} & \textbf{14}\%\cb{3.4}  & 10\%\cb{2.9} & 7\%\cb{2.5} \\

Hex & 25 & $(5\times5)$  & {83}\%\cb{3.7} & {56}\%\cb{4.9} & {85}\%\cb{3.5} & {60}\%\cb{4.8} & 79\%\cb{4.0} & {62}\%\cb{4.8}  & \textbf{86}\%\cb{3.4} & {65}\%\cb{4.7} \\
Hex & 49 & $(7\times7)$ & \textbf{76}\%\cb{4.2} & \textbf{49}\%\cb{4.9} & 44\%\cb{4.9} & 22\%\cb{4.1} & \textbf{57}\%\cb{4.9} & \textbf{34}\%\cb{4.7}  & 19\%\cb{3.9} & 8\%\cb{2.7} \\
Hex & 81 & $(9\times9)$ & \textbf{10}\%\cb{2.9} & {0}\%\cb{0.5} & 2\%\cb{1.3} & {0}\% & \textbf{4}\%\cb{1.8} & {0}\%\cb{0.5}  & {0}\%\cb{0.5} & 0\% \\

\multirow{1}{*}{Pentago} & 36 & $(6\times6)$  & 82\%\cb{3.8} & 41\%\cb{4.8} & \textbf{94}\%\cb{2.4} & \textbf{72}\%\cb{4.4} & 70\%\cb{4.5} & 28\%\cb{4.4}  & \textbf{80}\%\cb{4.0} & \textbf{46}\%\cb{4.9} \\

\multirow{1}{*}{Reversi} & 64 & $(8\times8)$  & {88}\%\cb{3.2} & {68}\%\cb{4.5} & {88}\%\cb{3.1} & {71}\%\cb{4.4} & \textbf{84}\%\cb{3.5} & \textbf{64}\%\cb{4.6}  & 63\%\cb{4.7} & 35\%\cb{4.6} \\

\multirow{1}{*}{The Mill Game} & 24 & $(8+8+8)$ & {67}\%\cb{4.3} & {36}\%\cb{4.1} & {70}\%\cb{4.2} & {40}\%\cb{4.3} & {61}\%\cb{3.9} & {34}\%\cb{3.6}  & {63}\%\cb{4.4} & {34}\%\cb{4.0} \\\hline
\multicolumn{3}{|c|}{Total average} & 75\% & 47\% & 69\% & 50\% & 64\% & 39\% & 56\% & 38\% \\\hline
\end{tabular}\end{center}
\caption{The results of the NN agents with 600 iterations. The baseline opponent is UCT with 600 ($1\times$) and 6,000 ($10\times$) iterations. The dataset for each game was gathered for at most 2h using 20 CPU cores. The 95\%-confidence intervals are given.
}\label{tab:main_results}
\end{table*}

\begin{table}[!ht]\small\centering\renewcommand{\arraystretch}{1.1}\setlength\tabcolsep{4.9pt}
\newcommand{\cb}[1]{{\scriptsize $\pm$#1}}%
\begin{tabular}{|l|r|r|r|}\hline
\multicolumn{1}{|c|}{\multirow{2}{*}{\bf Game}} & \multicolumn{3}{c|}{\bf Size of dataset (MCTS plays)} \\
                          & 200 & 400 & 1,000 \\\hline
                          & \multicolumn{3}{c|}{\bf Generation time + Training time} \\
Breakthrough $(6\times6)$ & 6s   + 20s &  11s + 20s & 25s + 20s \\
English Draughts          & 1.1m + 20s & 1.3m + 20s & 3.5m + 40s \\
Reversi                   & 1.2m + 20s & 1.2m + 20s & 3.5m + 40s \\\hline
                          & \multicolumn{3}{c|}{\bf Attention NN} \\
Breakthrough $(6\times6)$ & 4\%\cb{1.8} & 60\%\cb{4.8} & 74\%\cb{4.3} \\
English Draughts          & 31\%\cb{2.9} & 42\%\cb{3.2} & 54\%\cb{3.3} \\
Reversi                   & 25\%\cb{4.2} & 22\%\cb{4.0} & 28\%\cb{4.3} \\\hline
                          & \multicolumn{3}{c|}{\bf Convolutional NN} \\
Breakthrough $(6\times6)$ & 41\%\cb{4.8} & 48\%\cb{4.9} & 83\%\cb{3.7} \\
English Draughts          & 42\%\cb{3.4} & 55\%\cb{3.5} & 60\%\cb{3.4} \\
Reversi                   & 6\%\cb{2.3} & 7\%\cb{2.4} & 21\%\cb{4.0} \\\hline

\end{tabular}
\caption{Short training. The results of the NN agents with 600 iterations against UCT baseline with also 600 iterations.} \label{tab:fast}
\end{table}

\section{Our Method}

We adapted AlphaZero to the RBG framework and we focus on two-player zero-sum games, but the proposed method is limited only by the requirement of perfect information.
Our modifications focus on the following areas:

\textbf{Action space.}
The original AlphaZero approach requires the knowledge of action space, which must be either defined manually or inferred from the game rules (in GGP).
There are two issues.
First, the action space depends on a particular game encoding, and one game can have many implementations. Second, since tensors generally must have a fixed shape, games with a huge action space are particularly problematic, even if they have a small branching factor.
The second problem could be alternatively addressed by splitting actions into elementary fragments \cite{Kowalski2022SplitMoves}, yet this may lose heuristic information of actions.

To address these problems, we omit the policy network and attempt to use only the value network, abandoning any dependence on the action space. Thus, we use the standard UCT formula instead of the PUCT \cite{Rosin2011}, a variant of the UCT algorithm that uses a predictor to provide recommendations about the order of actions during exploration.

\textbf{Board topology.}
Another piece of information that must be extracted from game rules is the neighborhood of board tiles, which is necessary for the standard convolutional approach.
In GGP, extracting such information from the game description requires additional effort and is not always reliable. 
There is no guarantee that the games will have natural board topology, especially for non-rectangular topologies, or even the descriptions can be intentionally obfuscated.
Also, the natural board topology does not guarantee that adjacent cells are, by definition, the most correlated ones.

To avoid assumptions, we propose using an attention-based NN \cite{vaswani2017attention}. The model creates a (sinusoidal or learned positional) embedding for each tile and passes them through the encoder layers. This allows the network to learn the relations between tiles without any game-specific knowledge, regardless of the order of the tiles in the input. The presented method is the first application of attention networks to GGP. 
To evaluate the potential of self-attention, we propose comparing it with CNNs based on random permutations of the input board tiles. Reordering the tiles should eliminate the spatially local correlation, which is assumed by default by CNNs.

\textbf{Fast model generation}
To take advantage of the very fast RBG reasoner and to vastly decrease training time, the training dataset is generated by playing games using standard UCT MCTS agents (without NN). This approach provides better quality samples than in the early stages of self-play and allows gathering large amounts of data with limited resources. It can be easily parallelized since MCTS plays can be performed independently on CPU cores and do not require a GPU.

\section{Experiments with Conclusions}

We evaluated the trained NN agents by playing 400 games (200 per side) against a standard MCTS agent with game tree reuse. The results of the constant simulations limit are presented in Table~\ref{tab:main_results}.
Convolutional NN assumes that the board fits within a quad, and the tiles are given in order.
To examine the robustness, we tested the behavior after obfuscating by permuting the tiles randomly (the same permutation used for both NNs).
Overall, the attention NN has a similar performance to the CNN, even if the order of board tiles is random. The results vary a lot depending on the game, especially on its size.
Yet, they suggest the trend that attention is a better choice for larger games, which is particularly visible in size variations of Breakthrough and Hex.
It also suffers a smaller win ratio drop on average on our game set when the board is permuted (the mean drop is respectively 11\% and 8\% for attention NN vs. 13\% and 12\% for CNN).

Table~\ref{tab:fast} shows results after very short training.
We used smaller NNs here, trained on much less data (less than $3\%$ of the 2h dataset from Table~\ref{tab:main_results}).
This experiment was inspired by \cite{ThielscherAAAI20}, where in Breakthrough, after just 400 selfplay games, the winrate close to 80\% was achieved.
To achieve a similar playing strength, we needed about 1,000 MCTS vs.\ MCTS training plays. Thus, our plays are of worse quality, but we can generate them in less than a minute. 
Because of the RBG language efficiency \cite{Kowalski2020EfficientReasoning}, in our experiments, optimized Monte-Carlo playouts are much faster than NN evaluation. Such a situation, causing time-based comparison to favor pure MCTS, was not reported in GGP-related literature.

We conclude that knowledge of the action space and board topology is not required for obtaining good models.
The data generated from MCTS plays in place of self-plays is still useful; in some cases, it allows beating the baseline after a few minutes of computing.
Note that our research was focused on budget training, so the landscape could be different in longer settings.

\section{Acknowledgments}
This work was supported in part by the National Science Centre, Poland under project number 2021/41/B/ST6/03691.

\bibliography{bibliography}

\begin{thebibliography}{11}
\providecommand{\natexlab}[1]{#1}

\bibitem[{Browne et~al.(2012)Browne, Powley, Whitehouse, Lucas, Cowling,
  Rohlfshagen, Tavener, Perez, Samothrakis, and Colton}]{Browne2012ASurvey}
Browne, C.~B.; Powley, E.; Whitehouse, D.; Lucas, S.~M.; Cowling, P.~I.;
  Rohlfshagen, P.; Tavener, S.; Perez, D.; Samothrakis, S.; and Colton, S.
  2012.
\newblock {A Survey of Monte Carlo Tree Search Methods}.
\newblock \emph{IEEE Transactions on Computational Intelligence and AI in
  Games}, 4(1): 1--43.

\bibitem[{Cohen-Solal and Cazenave(2023)}]{cohen2023minimax}
Cohen-Solal, Q.; and Cazenave, T. 2023.
\newblock {Minimax Strikes Back}.
\newblock In \emph{AAMAS}, 1923--1931.

\bibitem[{Genesereth, Love, and Pell(2005)}]{Genesereth2005General}
Genesereth, M.; Love, N.; and Pell, B. 2005.
\newblock {General Game Playing: Overview of the AAAI Competition}.
\newblock \emph{AI Magazine}, 26: 62--72.

\bibitem[{Goldwaser and Thielscher(2020)}]{ThielscherAAAI20}
Goldwaser, A.; and Thielscher, M. 2020.
\newblock {Deep Reinforcement Learning for General Game Playing}.
\newblock \emph{AAAI}, 34(02): 1701--1708.

\bibitem[{Kowalski et~al.(2020)Kowalski, Miernik, Mika, Pawlik, Sutowicz,
  Szyku{\l}a, and Tkaczyk}]{Kowalski2020EfficientReasoning}
Kowalski, J.; Miernik, R.; Mika, M.; Pawlik, W.; Sutowicz, J.; Szyku{\l}a, M.;
  and Tkaczyk, A. 2020.
\newblock {Efficient Reasoning in Regular Boardgames}.
\newblock In \emph{IEEE Conference on Games}, 455--462.

\bibitem[{Kowalski et~al.(2022)Kowalski, Mika, Pawlik, Sutowicz, Szyku{\l}a,
  and Winands}]{Kowalski2022SplitMoves}
Kowalski, J.; Mika, M.; Pawlik, W.; Sutowicz, J.; Szyku{\l}a, M.; and Winands,
  M. H.~M. 2022.
\newblock {Split Moves for Monte-Carlo Tree Search}.
\newblock \emph{AAAI}, 36(9): 10247--10255.

\bibitem[{Kowalski et~al.(2019)Kowalski, Mika, Sutowicz, and
  Szyku{\l}a}]{Kowalski2019RegularBoardgames}
Kowalski, J.; Mika, M.; Sutowicz, J.; and Szyku{\l}a, M. 2019.
\newblock {Regular Boardgames}.
\newblock \emph{AAAI}, 33(1): 1699--1706.

\bibitem[{Rosin(2011)}]{Rosin2011}
Rosin, C.~D. 2011.
\newblock Multi-armed bandits with episode context.
\newblock \emph{Annals of Mathematics and Artificial Intelligence}, 61(3):
  203--230.

\bibitem[{Silver et~al.(2018)Silver, Hubert, Schrittwieser, Antonoglou, Lai,
  Guez, Lanctot, Sifre, Kumaran, Graepel et~al.}]{silver2018general}
Silver, D.; Hubert, T.; Schrittwieser, J.; Antonoglou, I.; Lai, M.; Guez, A.;
  Lanctot, M.; Sifre, L.; Kumaran, D.; Graepel, T.; et~al. 2018.
\newblock A general reinforcement learning algorithm that masters chess, shogi,
  and {G}o through self-play.
\newblock \emph{Science}, 362(6419): 1140--1144.

\bibitem[{Soemers et~al.(2021)Soemers, Mella, Browne, and
  Teytaud}]{soemers2021deep}
Soemers, D.~J.; Mella, V.; Browne, C.; and Teytaud, O. 2021.
\newblock Deep learning for general game playing with ludii and polygames.
\newblock \emph{ICGA Journal}, 43(3): 146--161.

\bibitem[{Vaswani et~al.(2017)Vaswani, Shazeer, Parmar, Uszkoreit, Jones,
  Gomez, Kaiser, and Polosukhin}]{vaswani2017attention}
Vaswani, A.; Shazeer, N.; Parmar, N.; Uszkoreit, J.; Jones, L.; Gomez, A.~N.;
  Kaiser, {\L}.; and Polosukhin, I. 2017.
\newblock Attention is all you need.
\newblock \emph{NeurIPS}, 30.

\end{thebibliography}
\end{document}